\documentclass[runningheads]{llncs}
\usepackage{graphicx}
\usepackage{amsmath}
\usepackage{amssymb}
\usepackage{booktabs}
\usepackage{algorithm}
\usepackage{algorithmic}
\usepackage{comment}
\usepackage{subfigure}
\usepackage{color}
\usepackage{bm}

\begin{document}
\title{Multimodal-GuideNet: Gaze-Probe Bidirectional Guidance in Obstetric Ultrasound Scanning}

\titlerunning{Multimodal-GuideNet: Gaze-Probe Bidirectional Guidance}
%
\author{
Qianhui Men\inst{1} \and
Clare Teng\inst{1} \and
Lior Drukker\inst{2, 3} \and
Aris T.\ Papageorghiou\inst{2} \and
J.\ Alison Noble\inst{1}
}
%
\authorrunning{
Q.\ Men et al.
}
%
\institute{
Institute of Biomedical Engineering,
\mbox{University of Oxford}, Oxford, UK\\
\email{qianhui.men@eng.ox.ac.uk}
\and
Nuffield Department of Women\textsc{\char13}s \& Reproductive Health, \mbox{University of Oxford, Oxford, UK}
\and
Department of Obstetrics and Gynecology, \mbox{Tel-Aviv University, Israel}
}
\maketitle              
\begin{abstract}
Eye trackers can provide visual guidance to sonographers during ultrasound (US) scanning. Such guidance is potentially valuable for less experienced operators to improve their scanning skills on how to manipulate the probe to achieve the desired plane. In this paper, a multimodal guidance approach (Multimodal-GuideNet) is proposed to capture the stepwise dependency between a real-world US video signal, synchronized gaze, and probe motion within a unified framework. To understand the causal relationship between gaze movement and probe motion, our model exploits multitask learning to jointly learn two related tasks: predicting gaze movements and probe signals that an experienced sonographer would perform in routine obstetric scanning. The two tasks are associated by a modality-aware spatial graph to detect the co-occurrence among the multi-modality inputs and share useful cross-modal information. Instead of a deterministic scanning path, Multimodal-GuideNet allows for scanning diversity by estimating the probability distribution of real scans. Experiments performed with three typical obstetric scanning examinations show that the new approach outperforms single-task learning for both probe motion guidance and gaze movement prediction. Multimodal-GuideNet also provides a visual guidance signal with an error rate of less than 10 pixels for a 224×288 US image.

\keywords{Probe Guidance \and Multimodal Representation
Learning \and Ultrasound Navigation \and Multitask Learning}
\end{abstract}
\section{Introduction}
Obstetric ultrasound (US) scanning is a highly-skilled medical examination that requires refined hand-eye coordination as the sonographer must look at a screen and simultaneously manipulate a handheld probe. Computer-assisted scanning with probe motion guidance could improve the training process for non-specialists to develop their scanning skills, which has been increasingly investigated among researchers and clinicians~\cite{housden2008calibration,prevost20183d,toporek2018autonomous}. Within the robotics field, work has focused on guiding operators to scan simple structures such as the liver~\cite{mustafa2013development}, lumbar and vertebrae~\cite{li2021autonomous}. Such solutions are not feasible for obstetric scans because of the variety of fetal anatomy to be measured and the unpredictable fetal movement.  

\begin{figure*}[t]
    \centering
    \includegraphics[width=0.9\textwidth]{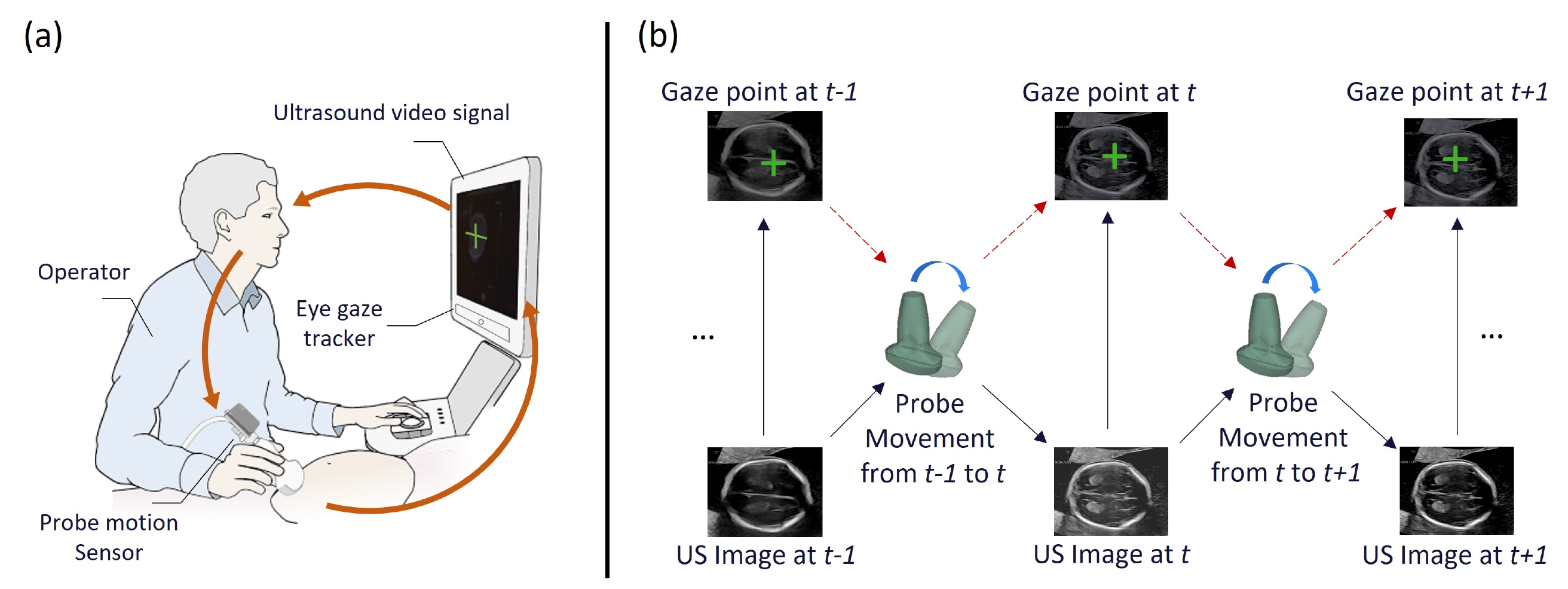}
    \caption{Data acquisition and the correspondence between captured signals. (a) Overview of the multi-modality data acquisition in the clinical obstetric ultrasound scanning. 
    (b) The unrolled US guiding process between the acquired US image, the probe motion signal, and the gaze signal.}
    \label{fig:motivation}
\end{figure*}

Previous studies in obstetric scanning guidance have proposed positioning the probe based on a behavioral cloning system~\cite{droste2020automatic} or landmark-based image retrieval~\cite{zhao2021visual}. In~\cite{droste2020automatic}, different strategies are modeled for operators to either follow the next-action instruction or directly approach the anatomical Standard Plane (SP)~\cite{baumgartner2017sononet}. Other work~\cite{toporek2018autonomous,wang2019robotic} deployed probe guidance signals to a robotic arm that is not practically applicable in a clinical environment. A common practice in these models is to treat probe guidance as an image-guided navigation problem. However, as multiple fetal anatomies can appear in a single US image, the gaze of the operator can provide instructive context about the likely next probe movement. Using gaze information to inform probe motion guidance has not been researched before now, and we explore this as the first aim of this work.

In addition to probe navigation, gaze information is also used as a guiding signal, usually in the form of gaze-point or saliency map (eye-tracking heat maps) prediction in US image or video. Cai et al.~\cite{cai2018multi,cai2018sonoeyenet} leveraged visual saliency as auxiliary information to aid abdominal circumference plane (ACP) detection, and Droste et al.~\cite{droste2019ultrasound} extended it to saliency prediction with diverse anatomical structures. Teng et al.~\cite{teng2021towards} characterized the visual scanning patterns from normalized time series scanpaths. Here, with the assumption that a sonographer will react to the next image inferred from their hand movement on probe, the second aim of this work is to explore whether probe motion is useful in guiding gaze.

In this work, we investigate how experienced sonographers coordinate their visual attention and hand movement during fetal SP acquisition. We propose the first model to provide useful guidance in both synchronized probe and gaze signals to achieve the desired anatomical plane. The model termed \textit{Multimodal-GuideNet} observes scanning patterns from a large number of real-world probe motion, gaze trajectory, and US videos collected from routine obstetric scanning (data acquisition in Fig.~\ref{fig:motivation}). \textit{Multimodal-GuideNet} employs multitask learning (MTL) for the two highly-related US guidance tasks of probe motion prediction and gaze trajectory prediction, and identifies commonalities and differences across these tasks. The performance boost over single-task learning models suggests that jointly learning gaze and probe motion leads to more objective guidance during US scanning. Moreover, the model generates real-time probabilistic predictions~\cite{graves2013generating} that provide unbiased guidance of the two signals to aid operators.

\section{Methods}
Figure~\ref{fig:motivation} outlines the principles of the approach. The probe orientation is recorded in 4D quaternions by an inertial measurement unit (IMU) motion sensor attached to the US probe, and the 2D gaze-point signal is captured by an eye-tracking sensor mounted on the bottom of the US screen. Given an US image starting at a random plane, its change in gaze between neighbour time steps, and its corresponding probe rotation, our multitask model \textit{Multimodal-GuideNet} estimates the instructive next-step movements of both the gaze and probe for the SP acquisition. The two tasks: probe motion prediction and gaze shift prediction complement each other for more accurate US scanning guidance. The problem definition and network architecture are as follows. 

\subsection{Problem Formulation}
Unlike previous US guidance models that only predict a fixed action, we regard the gaze and probe movements as random variables to account for inter- and intra-sonographer variation. For a more continuous prediction, the relative features are used from neighbour frames of these two modalities. Let $\bm{s}_t=\bm{g}_{t}-\bm{g}_{t-1}$ be the shift of gaze point $\bm{g}=(x,y)$ at time $t$ and $\bm{r}_t=\bm{q}_{t-1}^*\bm{q}_t$ be the rotation from the probe orientation $\bm{q}=(q_w,q_x,q_y,q_z)$, where $\bm{q}^*$ is the conjugate. We make the assumption that the gaze shift $\bm{s}_t$ follows a bi-variate Gaussian distribution, i.e., $\bm{s}_t\sim\mathcal{N}(\bm{\mu}_t^s, \bm{\sigma}_t^s, \rho_t^s)$, where $\bm{\mu}_t^s$ and $\bm{\sigma}_t^s$ denote the mean and standard deviation respectively in 2D, and $\rho_t^s$ is the correlation coefficient between $x$ and $y$. Therefore, at every step, the model outputs a 5D vector for gaze estimation. Similarly, we achieve a 14D vector for probe rotation $\bm{r}_t$ which follows a multi-variate Gaussian distribution $\bm{r}_t\sim\mathcal{N}(\bm{\mu}_t^r, \bm{\sigma}_t^r, \bm{\rho}_t^r)$. The multitask objective for model training is to jointly minimize the negative log-likelihoods of the two learning tasks
\begin{equation}
\mathcal{L} = \sum_{t=t_0}^T\Big(
-\lambda_s\log(\mathbb{P}(\bm{s}_t|\bm{\mu}_t^s, \bm{\sigma}_t^s, \rho_t^s))-\lambda_r\log(\mathbb{P}(\bm{r}_t|\bm{\mu}_t^r, \bm{\sigma}_t^r, \bm{\rho}_t^r))
+\eta (1 - \lVert \bm{\mu}_t^r\rVert^2)^2\Big),\label{eq: obj}
\end{equation} 
where $t_0$ and $T$ are the start and end indices for prediction. $\lambda_s$, and $\lambda_r$ control the training ratio of the two tasks and are both set to 1. $\eta$ is the weighting parameter for the quaternion prior to normalize $\bm{\mu}^r$ with $\eta=50$. 

\begin{figure*}[t]
    \centering
    \includegraphics[width=0.9\textwidth]{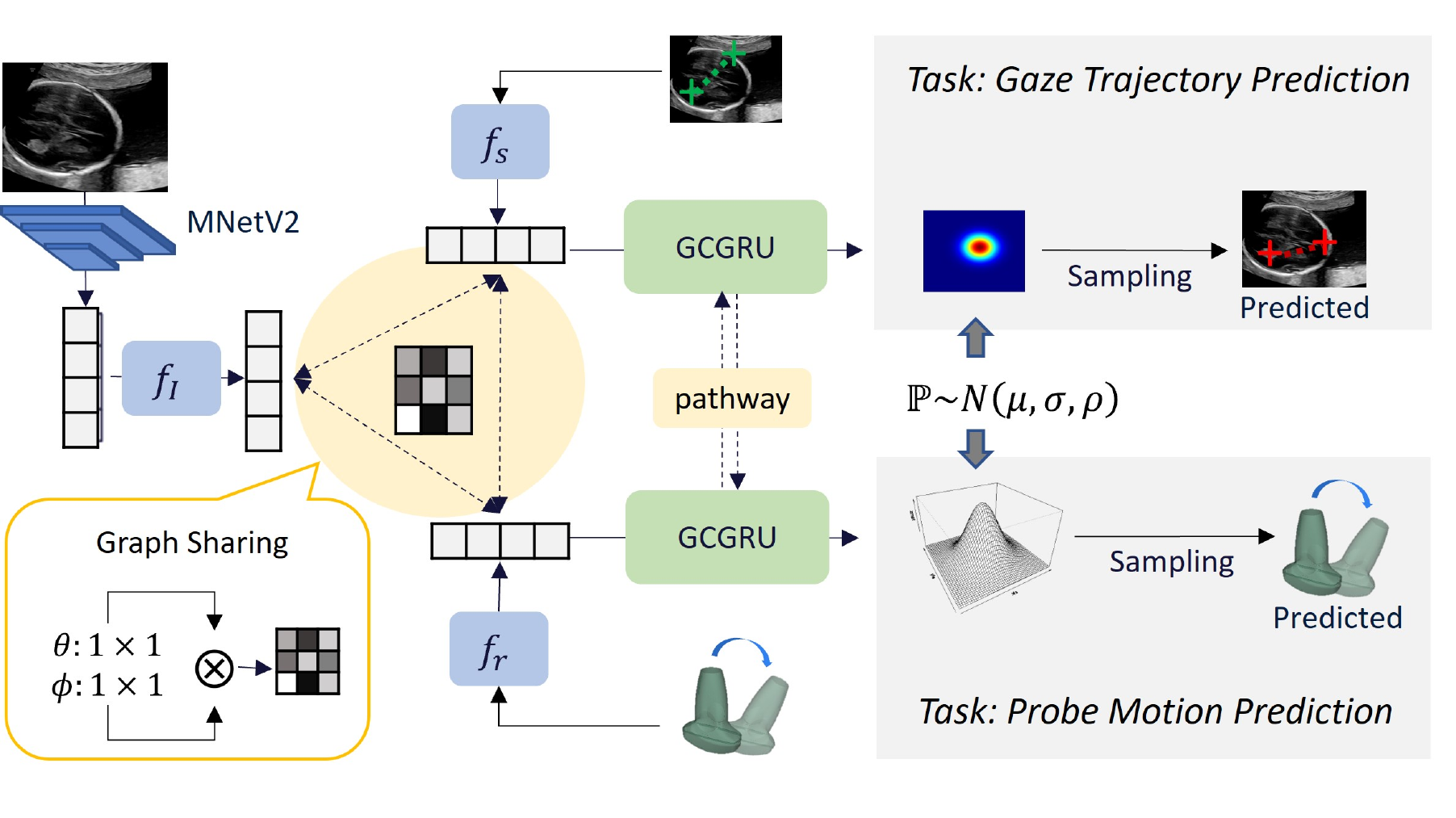}
    \caption{Flowchart of \textit{Multimodal-GuideNet} for a single time step. The two tasks share a modality-aware spatial graph from the three modalities.}
    \label{fig:framework}
\end{figure*}

\subsection{Multimodal-GuideNet}
To facilitate multitask learning, \textit{Multimodal-GuideNet} constructs a lightweight graph shared among the three modalities. The network backbone is formed by graph convolutional Gated Recurrent Unit (GCGRU)~\cite{li2016gated} that automatically allocates useful dependencies within the graph at each time step. The designed lightweight spatial graph is also computationally efficient for online inference. Temporally, the gaze and probe dynamics complement each other through a bidirectional pathway. The entire multitask framework is presented in Fig.~\ref{fig:framework}. To facilitate interactive learning within the graph structure, the input of each modality is embedded into an equal-sized 128-channel vector separately through a linear transformation block $f_I$, $f_s$, and $f_r$, each of which contains a fully-connected (FC) layer, a batch normalization (BN) layer, and a ReLU activation function. Before $f_I$, the grayscale US image is initially mapped to a flattened image representation $\bm{I}$~\cite{droste2020automatic} with MobileNetV2 (MNetV2)~\cite{sandler2018mobilenetv2}.

\textbf{Modality-aware Graph Representation Sharing.}
To model spatial proximity at time $t$, we propose a common graph structure $G_t=(\mathcal{V}_t, \mathcal{E}_t)$ that is shared among the three modalities, where $\mathcal{V}_t=\{f_I(\bm{I}_t), f_s(\bm{s}_t), f_r(\bm{r}_t)\}$ is the vertex set with 3 nodes. $\mathcal{E}_t$ is the edge set specified by a $3\times3$ adaptive adjacency matrix $A_t+M_t$ with the first term indicating the spatial relationship within $\mathcal{V}_t$, and the second term a trainable adjacency mask~\cite{yan2018spatial} to increase graph generalization. Inspired by~\cite{zhang2020semantics}, the edge weight of any two nodes in $A_t$ is formed by the affinity between the corresponding two modality features in the embedded space
\begin{equation}
    A_t(j,k)={\rm{softmax}}(\theta(f_j(j_t))^T\phi(f_k(k_t))), \quad j, k\in\{\bm{I}, \bm{s}, \bm{r}\}
\end{equation}
where $\theta$ and $\phi$ are $1\times1$ convolutions with $\theta(x),\phi(x)\in\mathbb{R}^{256}$, and the \textit{softmax} operation is to normalize the row summation of $A_t$. The message passed for $\bm{s}$ and $\bm{r}$ is therefore aggregated by one layer of a spatial graph convolution 
\begin{equation}
    \sum_{k\in\{\bm{I}, \bm{s}, \bm{r}\}}{\rm{sigmoid}}(A_t(j,k)f_k(k_t)W_j), \quad j\in\{\bm{s}, \bm{r}\}\label{eq:gcn}
\end{equation}
where $W_j$ is the input feature kernel specified for each gate in the GRU cell.

\textbf{Gaze-Probe Bidirectional Adaptive Learning.}
During US scanning, the gaze and probe movements of sonographers are generally heterogeneous, i.e., they do not move at the same pace. Upon approaching the SP, the gaze is more prone to rapid eye movements between anatomical structures while the probe remains steady. We account for this effect by enclosing a bidirectional inverse adaptive \textit{pathway} between hidden states of $\bm{s}$ and $\bm{r}$ in the time domain. Let $\bm{h}_t^s$, $\bm{h}_t^r$, $\tilde{\bm{h}}_t^s$, $\tilde{\bm{h}}_t^r$, and $\bm{z}_t^s$, $\bm{z}_t^r$ refer to the hidden state, candidate activation, and update gate of $\bm{s}$ and $\bm{r}$ in GRU at time $t$ respectively, we replace the original hidden state update $(1-\bm{z}_t)\odot \bm{h}_{t-1}+\bm{z}_t\odot \tilde{\bm{h}}_{t}$ with:
\begin{equation}
\begin{array}{l}
\bm{h}_{t}^{s}=\underbrace{\bm{\alpha}(1-\bm{z}_{t}^s)\odot\bm{h}_{t-1}^s+\bm{\alpha}\bm{z}_t^s\odot\tilde{\bm{h}}_t^s}_{\text{update from gaze}}+\underbrace{(1-\bm{\alpha})\bm{z}_{t}^r\odot \bm{h}_{t-1}^s+(1-\bm{\alpha})(1-\bm{z}_t^r)\odot\tilde{\bm{h}}_t^s}_{\text{inverse update from probe}}\\ 
\bm{h}_{t}^{r}=\underbrace{\bm{\beta}(1-\bm{z}_t^r)\odot\bm{h}_{t-1}^r+\bm{\beta}\bm{z}_t^r\odot\tilde{\bm{h}}_t^r}_{\text{update from probe}}+\underbrace{(1-\bm{\beta})\bm{z}_t^s\odot\bm{h}_{t-1}^r+(1-\bm{\beta})(1-\bm{z}_t^s)\odot\tilde{\bm{h}}_t^r}_{\text{inverse update from gaze}} \label{eq:pathway}
\end{array}
\end{equation}
where $\bm{\alpha}$, $\bm{\beta}$ are the adaptive channel-wise weights for $\bm{z}_t^s$ and $\bm{z}_t^r$, respectively, and $\odot$ is element-wise product. The number of hidden channels is set to 128 which is the same as $\bm{\alpha}$ and $\bm{\beta}$. With the proposed bidirectional pathway, the gaze and probe signals will adapt the domain-specific representation from each other to generate a more accurate scanning path. Other than the input operation for all gates (Eq.~\ref{eq:gcn}) and an adaptive hidden state (Eq.~\ref{eq:pathway}) for the output, we follow the operations in a standard GRU~\cite{cho2014learning} to transfer temporal information. 

\section{Experiments}
\subsection{Data}
The data used in this study were acquired from the PULSE (Perception Ultrasound by Learning Sonographic Experience) project~\cite{drukker2021transforming}. The clinical fetal ultrasound scans were conducted on a GE Voluson E8 scanner (General Electric, USA) and the video signal was collected lossless at 30 Hz. The corresponding gaze tracking data was simultaneously recorded as $(x,y)$ coordinates at 90 Hz with a Tobii Eye Tracker 4C (Tobii, Sweden). The probe motion was recorded with an IMU (x-io Technologies Ltd., UK) attached to the probe cable outlet as shown in Fig.~\ref{fig:motivation}(a). Approval from UK Research Ethics Committee was obtained for this study and written informed consent was also given by all participating sonographers and pregnant women. In total, there are 551 2\textsuperscript{nd} and 3\textsuperscript{rd} trimester scans carried out by 17 qualified sonographers. All three-modality data were downsampled to 6 Hz to reduce the time complexity.

\subsection{Experimental Settings}
The video frames were cropped to 224$\times$288 and irrelevant graphical user interface information was discarded. To facilitate image representation learning, we pre-train MNetV2 with a large number of cropped US frames under the 14 SonoNet standard plane classifier~\cite{baumgartner2017sononet} following the processing step of~\cite{droste2020automatic}. The clinical SP type is recognised automatically by Optical Character Recognition (OCR) with a total of 2121 eligible acquisitions labelled. For each acquisition, a multimodal data sample is selected 10s before the SP, which is the time for probe refinement. The raw gaze point is scaled to $(-0.5, 0.5)$ with the image center kept invariant, and the predicted $\bm{\mu}_t^s$ is also normalized to the same range by \textit{sigmoid} activation and a shift factor $0.5$ before the minimization of multitask objective. The ratio of train:test is 4:1. In the training stage, we randomly select 32 continuous frames in each sample. The model is evaluated for three biometry SPs which are trans-ventricular plane (TVP), abdominal circumference plane (ACP), and femur standard plane (FSP)~\cite{salomon2011practice}. The AdamW optimizer is adopted with an initial learning rate of 1e-3 decayed by 1e-2 every 8 epochs. The whole network is first trained on all 14 classes of SPs for 20 epochs and separately fine-tuned for TVP, ACP, and FSP for 16 epochs.

\subsection{Metrics and Baselines} 
We evaluate two probe scenarios: \textit{Coarse Adjustment} where probe rotation angle to SP $\geq$10\textdegree, and \textit{Fine Adjustment} $\leq$10\textdegree. The ratio of the two stages may vary from sample to sample and thus prediction performance is averaged among all frames in the same stage. For our method, we randomly sample 100 trajectories from the predicted distribution and average them as a final prediction $\hat{\bm{r}}$ and $\hat{\bm{s}}$. The two tasks are separately evaluated with different metrics: Probe movement is considered as correctly predicted if it is rotating towards the next target plane, i.e., $\angle(\bm{q}_{t-1}\hat{\bm{r}}_t, \bm{q}_{t})\leq\angle(\bm{q}_{t-1}, \bm{q}_{t})$; The predicted gaze point $\hat{\bm{g}}_t=\bm{g}_{t-1}+\hat{\bm{s}}_t$ is evaluated by pixel $l_2$ norm error. We compare our multitask model with two baselines and two single-task architectures: \textit{Baseline (r)}, continuing the previous probe rotation at the current time step; \textit{Baseline (g)}, using the previous gaze point at the current time step; \textit{US-GuideNet}~\cite{droste2020automatic}, single-task learning approach for probe guidance, where only probe motion is modeled and predicted from US video; \textit{Gaze-GuideNet}, single-task learning approach for gaze prediction, where only gaze information is modeled and predicted from US video by discarding the probe stream from Multimodal-GuideNet.

\begin{figure*}[t]
    \centering
    \includegraphics[width=\textwidth]{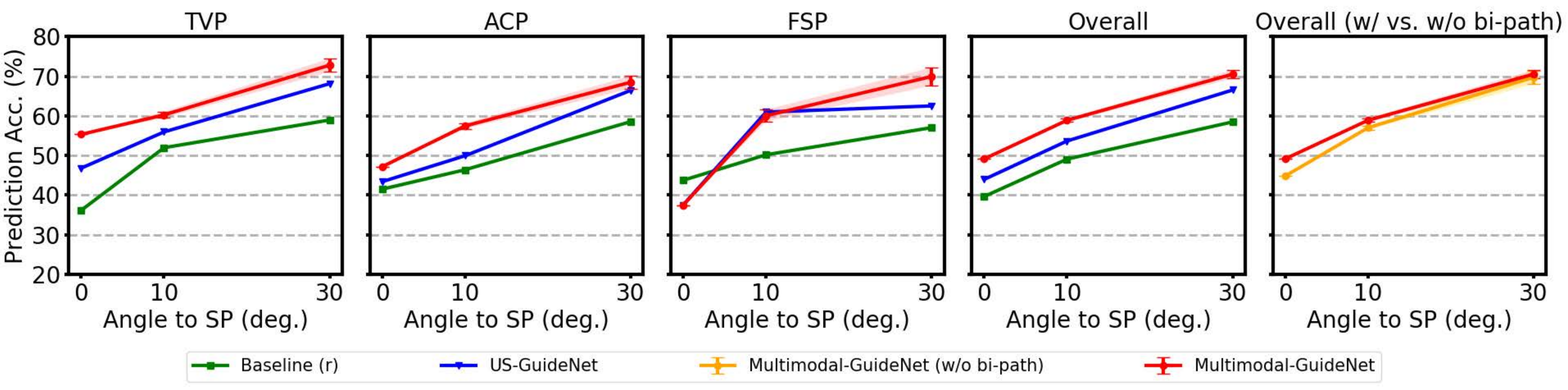}
    \caption{Probe rotation accuracy (the higher the better) on the 3 evaluated standard planes and the overall prediction with ablations. The shaded area indicates the standard deviation of our model across all 100 samplings. \textit{bi-path} signifies bidirectional pathway.}
    \label{fig:pred_acc}
\end{figure*}
\section{Results and Discussion}
\subsection{Probe Motion Guidance}
A detailed performance comparison for the probe guidance task is presented in Fig.~\ref{fig:pred_acc}. \textit{Multimodal-GuideNet} achieves an overall consistent improvement over the single task-based \textit{US-GuideNet}~\cite{droste2020automatic} for the two adjustment stages, which indicates that simultaneously learning the gaze patterns benefits the probe motion planning. The probe rotation for the femur (FSP) is difficult to predict when it gets close to the SP (at 0\textdegree). Different from a steady probe movement to achieve TVP and ACP, the probe manipulation close to FSP requires complicated twisting actions~\cite{salomon2011practice}. This also explains why incorporating gaze contributes more in the coarse adjustment (as at 30\textdegree) to locate the femur but not in the fine stage (as at 10\textdegree). Moreover, the flexible movements of fetal limbs increase the diversity in FSP detection, which explains why there is a slightly higher standard deviation observed for this plane. In the 5\textsuperscript{th} subplot (w/ vs. w/o bi-path), the improvements indicate that the pathway between gaze and probe stabilizes the probe movement with a more accurate prediction, especially in fine adjustment.

\begin{figure*}[b]
    \centering
    \includegraphics[width=\textwidth]{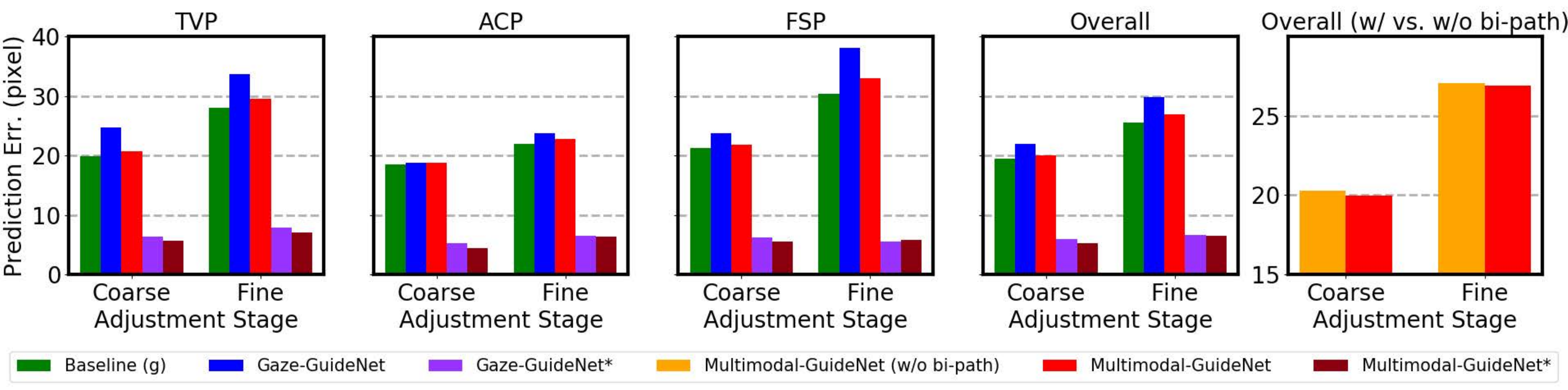}
    \caption{Gaze prediction error (the lower the better) on the 3 evaluated standard planes and the overall prediction with ablations. The error of the best-generated gaze point that is closest to ground truth is reported in \textit{Gaze-GuideNet*} and \textit{Multimodal-GuideNet*}, respectively.}
    \label{fig: gaze_err}
\end{figure*}

\subsection{Gaze Trajectory Prediction}
\begin{figure*}[t]
    \centering
    \includegraphics[width=\textwidth]{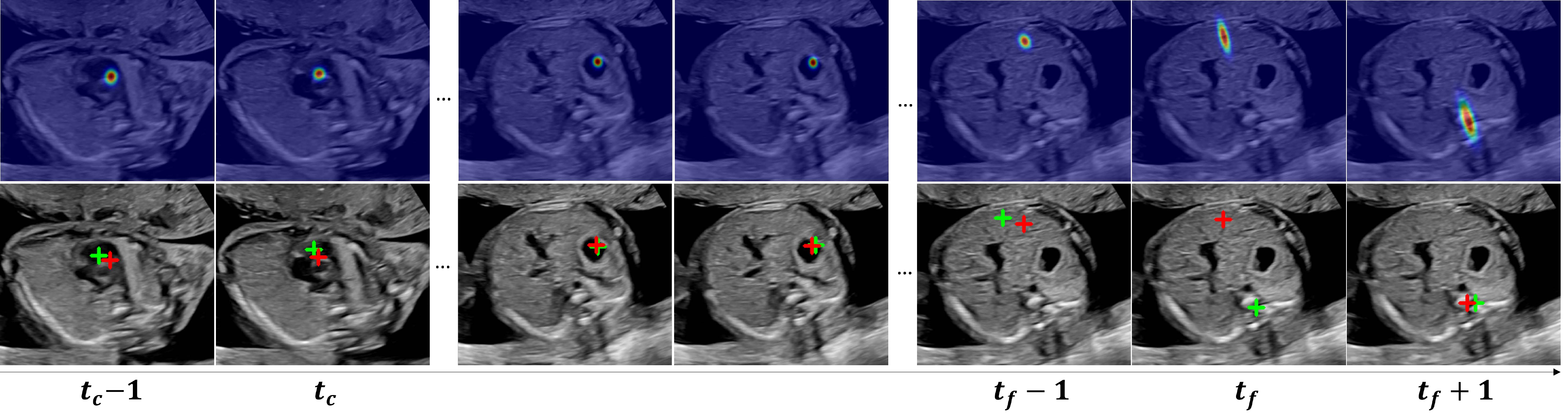}
    \caption{Visualization of predicted saliency map (top row), gaze point (bottom row, red star), and corresponding ground truth gaze point (green star) for an ACP searching sequence. $t_c$ and $t_f$ are timestamps for coarse and fine adjustment, respectively.}
    \label{fig:vis}
\end{figure*}
Figure~\ref{fig: gaze_err} shows the prediction results for the gaze task. A common observation among all three planes is that for gaze prediction, the error of fine adjustment is generally larger than coarse adjustment. This is because in contrast to the fine-grained probe motion, eye gaze movement during that time is quite rapid, flitting between the observed anatomical structures. When comparing the three planes, the error ranges are the lowest for ACP and the highest for FSP especially for the fine adjustment. Since the key anatomical structures in ACP are relatively close to each other, the sonographer requires a smaller change in gaze. For FSP, sonographers switch focus between both femur ends which increases the uncertainty of the gaze state in the next time step. Comparing between methods, \textit{Multimodal-GuideNet} reduces the error of \textit{Gaze-GuideNet} for all cases, which demonstrates the effectiveness of multitask learning over single-task learning in gaze prediction. The bidirectional pathway also slightly improves the gaze prediction as compared in the 5\textsuperscript{th} subplot. As a common evaluation in sampling-based generative models~\cite{gupta2018social}, we also report the performance of our best gaze point prediction among all samplings in \textit{Gaze-GuideNet*} and \textit{Multimodal-GuideNet*}. Their errors are within 10 pixels which shows the feasibility of the learned distribution in generating a plausible gaze trajectory. Practically, \textit{Multimodal-GuideNet*} could be useful when a precise gaze is needed such as when the sonographer focuses over a small range of underlying anatomical structure, and its improvement over \textit{Gaze-GuideNet*} indicates probe guidance could potentially help locate such a fixation point.

Figure~\ref{fig:vis} shows an example of predicted visual saliency and gaze point deduced from the generated gaze shift distribution. The predictions are highly accurate in all timestamps except for a significant gaze shift at frame $t_f$. However, the predicted saliency map at $t_f$ correctly estimates the orientation of gaze shift. Saliency map-based numerical metrics are also evaluated in the supplementary material, where the multitask model generally outperforms the single-task one. In general, modeling the gaze information as a bi-variate distribution is technically advantageous over a saliency map-based predictor, as the problem complexity is reduced from optimizing a large feature map to only a few parameters for probability density estimation. The flexibility in gaze sampling also preserves the variety in gaze movements.

\section{Conclusion}
We have presented a novel multimodal framework for bidirectional guidance between probe motion and eye-tracking data in routine US scanning. We have explored multitask learning by jointly predicting the probe rotation and gaze trajectory from US video via a shared modality-aware graph structure. The performance gains over single-task predictions suggest that the two-modality signals complement each other to reach the scanning target, while ignoring any of them will lead to a biased guidance. The learned guidance signals with probability distributions also allow for diversity between individual scans in a practical environment. 

\section*{Acknowledgements}
We acknowledge the ERC (ERC-ADG-2015 694581, project PULSE), the EPSRC (EP/MO13774/1, EP/R013853/1), and the NIHR Oxford Biomedical Research Centre.

%
%
%
\bibliographystyle{splncs04}
\bibliography{ref}

\end{document}